%% file: main.tex
\pdfoutput=1
\documentclass[11pt]{article}

\usepackage[preprint]{acl}

\usepackage{times}
\usepackage{latexsym}

\usepackage[T1]{fontenc}
\usepackage[utf8]{inputenc}

\usepackage{microtype}

\usepackage{inconsolata}

\usepackage{graphicx}

\usepackage{booktabs}
\usepackage{enumitem}
\usepackage{algorithm}
\usepackage{algpseudocode}
\usepackage{amsmath} 
\usepackage{multirow}
\usepackage{subcaption}
\usepackage{lipsum} 
\usepackage{float}

\definecolor{myblue}{HTML}{2069d6}
\definecolor{myred}{HTML}{c74d46}
\definecolor{mypurple}{HTML}{893aba}
\newcommand{\vivid}[1]{\textcolor{myblue}{\textbf{\textit{#1}}}} 
\newcommand{\technical}[1]{\textcolor{myred}{\textbf{\textit{#1}}}} 
\newcommand{\note}[1]{\textcolor{mypurple}{\textbf{\textit{#1}}}}

\title{LLM-Collaboration on  Automatic Science Journalism \\ for the General Audience}

\author{Gongyao Jiang \and Xinran Shi \\
  The Hong Kong University of \\ Science and Technology (Guangzhou)\\
  \texttt{jianggongyao@gmail.com} \\
  \And
  Qiong Luo\thanks{Corresponding Author} \\
  The Hong Kong University of \\Science and Technology (Guangzhou) / \\
  The Hong Kong University of \\Science and Technology \\
  \texttt{luo@cse.ust.hk} \\}

\begin{document}
\maketitle
\begin{abstract}
Science journalism reports current scientific discoveries to non-specialists, aiming to enable public comprehension of the state of the art.
However, this task can be challenging as the audience often lacks specific knowledge about the presented research.
To address this challenge, we propose a framework that integrates three LLMs mimicking the real-world writing-reading-feedback-revision workflow, with one LLM acting as the journalist, a smaller LLM as the general public reader, and the third LLM as an editor.
The journalist's writing is iteratively refined by feedback from the reader and suggestions from the editor.
Our experiments demonstrate that by leveraging the collaboration of two 7B and one 1.8B open-source LLMs, we can generate articles that are more accessible than those generated by existing methods, including advanced models such as GPT-4. 
% Our code will be publicly available for research purposes.

\end{abstract}

\input{src/1-intro}

\input{src/3-methd}
\input{src/4-exprt}

\input{src/5-discs}
\input{src/2-relat}
\input{src/6-concl}

% \section*{Acknowledgments}

%\bibliography{anthology, ,custom}
\bibliography{custom}

\appendix

\input{src/appdx}

\end{document}

%% file: src/1-intro.tex
\section{Introduction}
Science journalism creates journalistic content that covers a wide range of scientific research, enhancing the public's understanding of science \citep{gopfert2008strength, allan2011introduction, angler2017science}. 
However, with rapid advances in various disciplines, science journalism struggles to keep pace with the exponential growth of knowledge. 
In response, automatic science journalism (ASJ) has been proposed to expedite the filtering, learning, and communication of scientific knowledge \citep{dangovski2021we}.

\begin{figure}
    \centering
    \includegraphics[width=0.5\textwidth]{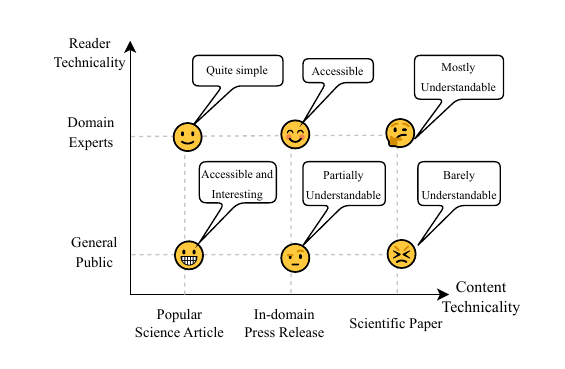}
    \caption{Reader experience varies with content technicality. Science journalism for the general audience demands high accessibility.}
    \label{fig:com}
\end{figure}

The essence of ASJ lies in elucidating complex technical content for readers, thereby facilitating their comprehension of advanced research \citep{cardenas2023don}. 
However, ASJ-generated content can be challenging for the general audience who lack in-depth knowledge of specific fields. 
As depicted in Figure \ref{fig:com}, the degree to which content is embraced varies among readers with different levels of domain knowledge \citep{august2024know}, underscoring the need for high readability to broaden a diverse readership.
Some researchers have developed parallel corpora \citep{dangovski2021we, goldsack2022making, cardenas2023don}, where the target content is extracted from online scientific news or journals. 
However, these press releases often remain technical, likely because they are originally tailored for professional researchers rather than the general audience. 
Models trained on such content struggle to generate materials easily understandable for a broader audience.

\begin{figure*}[t]
    \centering
    \includegraphics[width=1\textwidth]{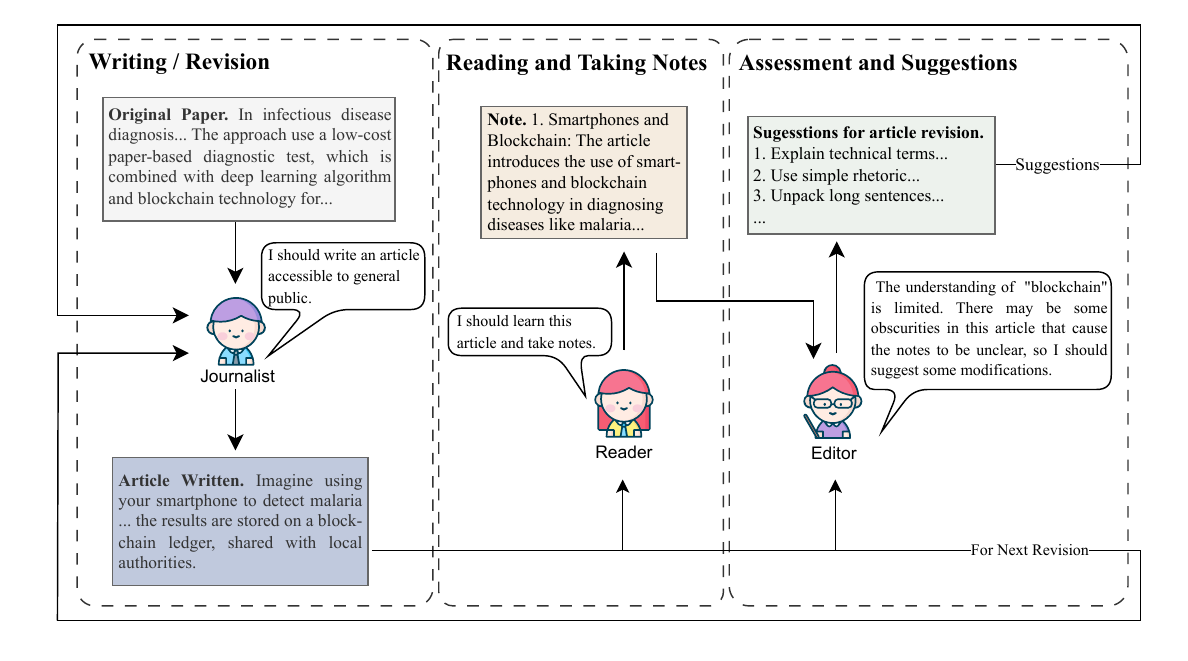}
    \caption{Overview of our LLM-Collaboration framework, LLM-CLBR.}
    \label{fig:overview}
\end{figure*}

Large language models (LLMs) have shown impressive proficiency in instruction adherence and content generation \citep{openai2024gpt4, bai2023qwen}, thereby making them potential tools for ASJ.
Furthermore, LLMs have exhibited social intelligence \citep{park2023generative}, enabling them to play realistic roles and collaborate in real-world tasks \citep{qian2023communicative, talebirad2023multi}.
Motivated by these observations, we propose a novel framework that leverages LLMs as communicative agents to collectively accomplish the ASJ task.

Our goal is to automatically generate a popular science article based on a technical paper and make it accessible to the general public. In the real world, a journalist typically receives revision suggestions from a professional editor; to make it accessible to the general public, the journalist could pass their article to friends without domain expertise on the topic to get feedback for revision.  Therefore, we design our framework to simulate such real-world scenarios, in which three LLMs collaborate to go through a four-step iterative process, which includes writing, reading, feedback, and revision, to generate highly accessible popular science articles. The questions for our framework are then (1) how the reader gives feedback; (2) how the editor gives suggestions based on the reader feedback; (3) how the journalist takes suggestions to revise the generated article; and (4) whether this iterative process improves the quality, especially accessibility, of the generated article.

As depicted in Figure \ref{fig:overview}, we have an LLM serve as the journalist writing for readers who lack domain knowledge of the given paper. We have another LLM, smaller than the journalist LLM, which acts as a general reader, to read the generated article and take notes to give reading feedback. 
As a less proficient model, the reader LLM needs material that is easily understandable to take comprehensive notes on.
Therefore, the more accessible the explanations in the written article are, the greater the clarity and accuracy of the notes will exhibit.

LLMs have shown the capability of evaluating the quality of text \citep{chan2023chateval, zheng2024judging, desmond2024evalullm}.
Therefore, we let an editor (the third LLM) evaluate the correctness and comprehensiveness of the reader's notes and then provide suggestions for the revision of the journalist's article. 
The journalist then revises the previous version of the article based on the suggestions. 
By this iterative and tuning-free process, the popular science article is enhanced continuously and made more accessible to a general audience.
To the best of our knowledge, our work is the first comprehensive study on LLMs for ASJ.

To assess the proposed method for ASJ, we employ both automatic metrics and human evaluation on measures including readability, information conveyance, authenticity, and interestingness of our generated articles. Compared with other methods, including those with fine-tuning and prompting on various LLMs, our proposed method achieves the highest readability while remaining competitive on the other measures. 
We also provide a detailed analysis, including ablation studies of removing the editor LLM, removing the reader LLM, or removing both, as well as trend analysis and case studies, to offer a comprehensive understanding of LLMs in the ASJ task.

In brief, we make the following contributions: 
\begin{itemize}[noitemsep, topsep=0pt] 
    \item A novel ASJ framework with collaborative LLMs, generating articles of high readability. 
    \item Comprehensive experiments, analyses, and recommendations for LLM usage in ASJ. 
\end{itemize}

%% file: src/3-methd.tex
\section{Methodology}
Following \citet{dangovski2021we, goldsack2022making, cardenas2023don}, ASJ aims to automatically distill a scientific paper into an article accessible to a broader audience.
Our ASJ framework employs an iterative workflow of writing, reading, suggestion-making, and revision, as illustrated in Figure \ref{fig:overview}.
All prompts for each LLM agent are listed in Appendix \ref{sec:prompts}.

\subsection{The LLM Journalist}
LLMs have shown strong writing abilities \citep{yuan2022wordcraft, wasi2024llms}. 
Thus, they are promising tools for rewriting a provided paper into a more accessible version.
Following established strategies \citep{zheng2024judging, zhang2024llm}, we start with prompting an LLM to assume the role of a journalist.
Subsequently, the LLM is prompted that, given the paper, its task is to compose an article for the general public.

\subsection{The LLM Reader}

\begin{figure}[t]
    \centering
    \includegraphics[width=0.485\textwidth]{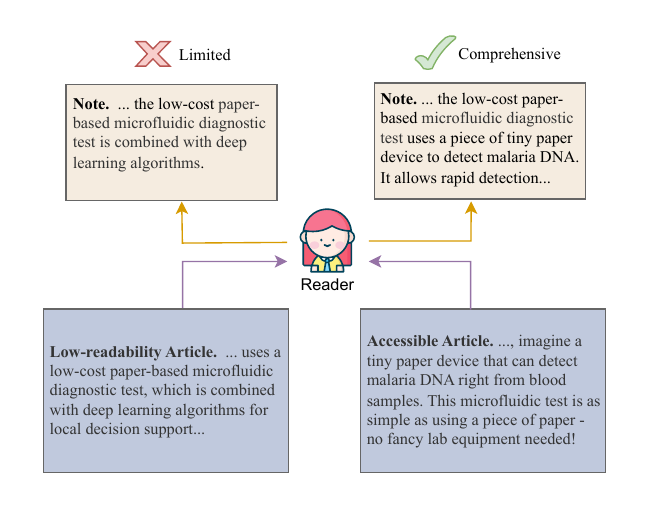}
    \caption{Accessible content helps the reader take comprehensive notes.}
    \label{fig:student}
\end{figure}

In our preliminary attempts, we asked the journalist LLM to directly assess the readability of the content. However, the results are unsatisfactory,  probably due to the gap between human and model perceptions of reading difficulty. 
As illustrated in the two text boxes at the bottom of Figure \ref{fig:student}, LLMs may regard both pieces of writing are of a similar level of readability, as they all incorporate essential information, even if the terminology ``low-cost paper-based microfluidic diagnostic tests'' on the left side is not clearly explained.
However, in the eyes of a human reader, the content on the right is perceived as more accessible.\footnote{We briefly document other failed attempts in Appendix \ref{sec:fail} for reader information.} 

To address this readability assessment problem, we design a separate reader LLM to read the content and generate reading notes. Our idea is inspired by the accumulation of errors, a common phenomenon in pipeline systems \citep{caselli2015s, wu2018beyond, jiang2023pilot, dziri2024faith}. Specifically, we can utilize the propagation from the textual readability of the journalist's article to the reading comprehensibility of the reader in the writing-reading pipeline, to induce the readability of the generated article to become explicit.

Different from the LLM journalist, the reader LLM is of a smaller scale and simulates a general reader with limited domain knowledge. 
Once presented with the article crafted by the journalist, the reader LLM is employed to read the article and take notes.
Specifically, we instruct the reader LLM to explain key terms in the article by extracting the explanations, if present, directly from the article or offering explanations for these terms, otherwise.

Intuitively, if the article is more accessible, the reading notes will be more comprehensive. 
For instance, in Figure \ref{fig:student}, if the piece (left) lacks a detailed explanation of the term ``paper-based microfluidic diagnostic test'', the reader will only note this term as being ``combined with deep learning''. 
If the article (right) explains the usage and advantages of this term in plain and readable language, the reader LLM can grasp this knowledge.
Through this readability propagation from the journalist's article to the comprehensiveness of the reader's notes, the editor LLM can better recognize issues in the journalist's writing and then provide suitable suggestions for modifications.

\subsection{Automated Suggestions and Revisions}
LLMs have demonstrated strong capabilities in serving as evaluators, widely utilized in various generative tasks \citep{chan2023chateval, zheng2024judging, desmond2024evalullm}.
Therefore, we employ an LLM as a senior editor for automated evaluation of reader comprehension and providing recommendations for article enhancement.
Given the article from the journalist and notes from the reader, an LLM editor is tasked with assessing the quality of the reader's notes and identifying issues in the journalist's writing, which may lead to reading obstacles. 

Next, the editor offers advice for the journalist's content development.
For example, in Figure \ref{fig:overview}, the editor finds that the reader's understanding of the term ``blockchain'' is limited, possibly due to an insufficient explanation in the reading material. 
To address this perceived issue, the editor suggests that the article should ``explain technical terms''. 
These suggestions are then incorporated into the instructions that will guide the journalist in revising the article.

Subsequently, with the strong ability to follow instructions, the journalist LLM rewrites the article according to the suggestions.
Then, the revised piece is fed to the reader for reading and taking notes to continue the process. 
By an iterative cycle encompassing writing, note-taking, suggesting modifications, and revision among three LLMs, the article tailored for the general readership undergoes steady enhancement.
We further analyze this process in Section \ref{sec:ana}.

%% file: src/4-exprt.tex
\section{Experiments}

\subsection{Settings}
\noindent \textbf{Datasets.}
We use three publicly available corpora in different disciplines as benchmarks, namely \emph{SCITech}, \emph{eLife}, and \emph{PLOS}.
% Appendix \ref{sec:hyper} presents a brief introduction of these datasets.
% Table \ref{tab:stat} shows some statistics of these three datasets.
Appendix \ref{sec:hyper} presents a brief introduction and some statistics of these datasets.
The same as the previous studies, in \emph{SCITech}, we use 1431 instances for training and validation and the remaining 1000 for testing.
We also separate each of \emph{eLife} and \emph{PLOS} datasets into training, validation, and testing splits at a ratio of 90\%/5\%/5\%.
% \begin{table}[t]
%     \small
%     \centering
%     \begin{tabular}{lrrrrr}
%         \toprule
%        Statistic & \textbf{SCITech} & \textbf{eLife} & \textbf{PLOS} \\
%         \midrule
%         \# $\text{pairs}$ & 2431 & 4828 & 27525 \\
%         \# $\text{words}^\text{ori}$ & 216.8 & 166.3 & 268.3\\
%         \# $\text{sentences}^\text{ori}$ & 5.7 & 6.8 & 10.2 \\
%         \# $\text{words}^\text{pln}$ & 176.1 & 347.6 & 175.6 \\
%         \# $\text{sentences}^\text{pln}$ & 7.9 & 15.7 & 7.8\\
%         \bottomrule
%     \end{tabular}
%     \caption{Statistics of benchmark datasets. Words and sentences are average values. The ``ori'' superscript indicates abstracts of original papers, and ``pln'' represents plain summaries written by authors or journalists.}
%     \label{tab:stat}
% \end{table}

\begin{table*}[t]
\centering
\small
\begin{tabular}{lcccccccccc}
\toprule
                                               & \multicolumn{3}{c}{\textbf{SCITech}}                                                                                 & \multicolumn{3}{c}{\textbf{eLife}}                                                                                  & \multicolumn{3}{c}{\textbf{PLOS}}                                                                                    &                                       \\ \cmidrule(lr){2-4} \cmidrule(lr){5-7} \cmidrule(lr){8-10}
\multirow{-2}{*}{\textbf{Approach}}            & \textbf{CLI}$\downarrow$                        & \textbf{FKGL}$\downarrow$                      & \textbf{DCRS}$\downarrow$                       & \textbf{CLI}$\downarrow$                    & \textbf{FKGL}$\downarrow$                   & \textbf{DCRS}$\downarrow$                    & \textbf{CLI}$\downarrow$                    & \textbf{FKGL}$\downarrow$                    & \textbf{DCRS}$\downarrow$                 & \multirow{-2}{*}{\textbf{Avg.}}       \\ 
\midrule
{\color[HTML]{485aa3} \textbf{LLaMA-2-7B}}     & 15.13                                 & 13.79                                 & 10.38                                & 15.16                                 & 14.03                                & 10.50                                & 15.36                                 & 14.28                                 & 10.54                                & 13.24                                 \\
{\color[HTML]{485aa3} \textbf{Gemma-7B}}       & 14.93                                 & 13.75                                 & 10.52                                & 15.01                                 & 12.08                                & 11.03                                & 15.52                                 & 12.29                                 & 10.92                                & 12.89                                 \\
{\color[HTML]{485aa3} \textbf{Mistral-7B}}     & 14.90                                 & 13.54                                 & 10.82                                & 14.61                                 & 11.72                                & 10.85                                & 15.38                                 & 11.98                                 & 11.21                                & 12.78                                 \\
{\color[HTML]{485aa3} \textbf{Qwen-1.5-7B}}    & 14.77                                 & 13.50                                 & 10.72                                & 14.72                                 & 11.83                                & 10.92                                & 15.06                                 & 11.94                                 & 11.09                                & 12.73                                 \\
{\color[HTML]{485aa3} \textbf{LLaMA-3-8B}}     & 14.84                                 & 13.18                                 & 10.41                                & 14.55                                 & 11.65                                & 10.49                                & 15.18                                 & 12.01                                 & 10.88                                & 12.58                                 \\
{\color[HTML]{485aa3} \textbf{Mixtral-8x7B}}   & 13.98                                 & 13.25                                 & 10.36                                & 14.21                                 & 12.01                                & 10.28                                & 15.34                                 & 11.58                                 & 10.98                                & 12.44                                 \\
{\color[HTML]{485aa3} \textbf{Qwen-1.5-72B}}   & 13.78                                 & 13.10                                 & 10.25                                & 14.17                                 & 12.09                                & 10.35                                & 15.18                                 & 11.75                                 & 10.62                                & 12.37                                 \\
{\color[HTML]{48968B} \textbf{GPT-3.5-Turbo}}  & 14.98                                 & 13.62                                 & 10.81                                & 14.35                                 & 11.87                                & 10.98                                & 15.11                                 & 11.92                                 & 10.87                                & 12.72                                 \\
{\color[HTML]{48968B} \textbf{GPT-4}}          & 13.48                                 & 12.13                                 & 10.14                                & 13.96                                 & 10.87                                & 10.11                                & 14.86                                 & 11.78                                 & 10.47                                & 11.98                                 \\
{\color[HTML]{d17630} \textbf{BART-FT}}           & 13.43                                 & 15.22                                 & 10.66                                & 12.32                                 & 10.65                                & 9.19                                 & 15.61                                 & 14.24                                 & 10.51                                & 12.43                                 \\
{\color[HTML]{d17630} \textbf{Qwen-1.5-7B-FT}} & 13.37                                 & 14.79                                 & 10.48                                & 12.15                                 & 10.63                                & \underline{9.12}                                 & 15.54                                 & 13.95                                 & 10.58                                & 12.29                                 \\
{\color[HTML]{754494} \textbf{LLM-WS-CLBR}}   & 12.94                                 & 13.33                                 & 10.33                                & 12.04                                 & {\textbf{9.85}} & {\textbf{9.04}} & 13.15                                 & 11.48                                 & 10.17                                & 11.37                                 \\
{\color[HTML]{754494} \textbf{LLM-CLBR}}      & {\textbf{12.69}} & {\textbf{10.16}} & {\underline{9.79}} & {\textbf{11.60}} & 10.10                                & 9.46                                 & {\underline{12.74}} & {\underline{10.00}} & {\textbf{9.69}} & {\textbf{10.69}} \\
\hline
Reader: 1.8B $\rightarrow$ 7B  & \underline{12.81} & \underline{10.35} & \textbf{9.68} & \underline{11.82} & \underline{10.01} & 9.51 & \textbf{12.67} & \textbf{9.93} & \underline{9.78} & \underline{10.73} \\
$-$ Reading Notes	& 13.21 & 10.63 & 10.33 & 12.22 & 10.78 & 10.02 & 13.35 & 10.59 & 10.25 & 11.26 \\
$-$ Suggestions	& 13.25 & 10.69 & 10.39 & 12.17 & 10.83 & 10.08 & 13.31 & 10.74 & 10.42 & 11.32 \\
$-$ Collaboration	& 13.50 & 11.01 & 10.71 & 12.47 & 10.99 & 10.41 & 13.65 & 10.91 & 10.70 & 11.59 \\
\midrule
Paper Abstracts                            & 16.67                                 & 15.27                                 & 11.39                                & 17.53                                 & 15.35                                & 11.87                                & 16.38                                 & 14.98                                 & 11.10                                & 14.50                                 \\
Plain Summaries                                  & 14.23                                 & 14.79                                 & 11.13                                & 12.52                                 & 10.91                                & 8.94                                 & 15.90                                 & 14.76                                 & 10.91                                & 12.68                                 \\ 
\bottomrule
\end{tabular}
\caption{The results of automated evaluation. We tested various methods, including {\textbf{\color[HTML]{485aa3}the open-source LLM prompting}}, {\color[HTML]{48968B}\textbf{the closed LLM prompting}}, {\color[HTML]{d17630}\textbf{fine-tuning}}, and {\color[HTML]{754494}\textbf{collaboration of LLMs (ours)}}.}
\label{tab:auto}
\end{table*}

\begin{table}[t]
\centering
\small
\begin{tabular}{lcccc}
\toprule
\textbf{Approach} & \textbf{Read.} & \textbf{Info.} & \textbf{Auth.} & \textbf{Intr.} \\ \hline
\multicolumn{5}{c}{\textbf{Within Field}}                                                                              \\ \hline
Plain   Summaries & 2.95          & 2.90          & 3.35          & 2.70          \\
Qwen1.5-7B      & 3.50          & 3.35          & 3.40          & 3.10          \\
GPT-4           & 3.80          & \textbf{3.75} & \textbf{3.80} & 3.40          \\
LLM-CLBR        & \textbf{3.95} & 3.60          & 3.70          & \textbf{3.55} \\
\hline
\multicolumn{5}{c}{\textbf{Outside Field}}                                                 \\ \hline
Plain   Summaries & 2.75          & 2.85          & 3.25          & 2.65          \\
Qwen1.5-7B      & 3.35          & 3.10          & 3.30          & 3.10          \\
GPT-4           & 3.40          & \textbf{3.55} & \textbf{3.70} & 3.15          \\
LLM-CLBR        & \textbf{3.65} & 3.40          & 3.55          & \textbf{3.20}  \\ \bottomrule
\end{tabular}
\caption{Results of human evaluation, where `Read.' indicates `Readability,' `Info.' denotes `Information Conveyance,' `Auth.' represents `Authenticity,' and `Intr.' signifies `Interestingness'.}
\label{tab:human}
\end{table}

\noindent \textbf{Methods For Comparison.}
\begin{itemize}[noitemsep, topsep=1pt]
    \item \textbf{BART.} \citet{goldsack2022making, cardenas2023don} used BART \citep{LewisBart} for ASJ, showing strong performance.
    \item \textbf{LLMs.} We test the performance of various LLMs, including both open-source and closed LLMs, i.e., LLaMA-2-7B \citep{touvron2023llama}, Gemma-7B \citep{gemma}, Mistral (7B, 8x7B, \citealp{jiang2023mistral}), Qwen1.5 (7B, 72B, \citealp{bai2023qwen}), LLaMA-3-8B \citep{llama3}, GPT-3.5-Turbo-1106 \citep{openaichatgpt}, and GPT-4-1106-preview \citep{openai2024gpt4}. We prompt these LLMs and also fine-tune the Qwen1.5-7B model for ASJ.
    \item \textbf{Our Methods.} We test two versions of collaborating LLMs. One is 2$\times$Qwen1.5-7B+Qwen1.5-1.8B (LLM-CLBR, CLBR for `collaboration'). Another is this combination where the journalist is replaced by a fine-tuned version as a warm start (LLM-WS-CLBR).
\end{itemize}

\noindent \textbf{Automatic Evaluation.}
Following \citet{goldsack2022making, cardenas2023don}, we use Coleman-Liau Index (CLI), Flesch-Kincaid Grade Level (FKGL) and Dale-Chall Readability Score (DCRS) to automatically assess the readability.
CLI considers the count of sentences, words, and characters, while FKGL is based on the number of sentences, words, and syllables. 
DCRS assesses readability by analyzing the average sentence length and the presence of familiar words from a list of the most commonly used words.

\noindent \textbf{Human Evaluation.}
Automatically assessing the authenticity and informativeness of content has been a challenging task.
\citet{cardenas2023don} used QuestEval \citep{Scialom_Dray_Lamprier_Piwowarski_Staiano_Wang_Gallinari_2021} to assess the faithfulness of ASJ-generated content, yet the results exhibited significant variances.
Therefore, human evaluation remains the main method for such assessments.
We enlist four human participants for evaluation. 
All of them are including master's students or holders of master's degrees, two from computer science and two from biomedical science.
Specifically, we sample 10 pairs of original papers and generated articles in computer science from SCITech, as well as 10 pairs in biomedical science pairs, 5 pairs from eLife and 5 pairs from PLOS.

We choose four representative methods for human evaluation: (1) plain summaries by human writers, (2) Qwen1.5-7B generation, (3) GPT-4 generation, and (4) generation by our LLM-Collaboration method.
The human evaluation encompasses multiple dimensions, namely Readability, Information Conveyance, Authenticity, and Interestingness.
Participants are tasked with evaluating the articles using a 1-5 Likert scale \citep{likert1932technique}, grounded on specific questions.
Each participant is assigned to assess all articles both in the field they are familiar with and those they are not familiar with, to provide a genuine evaluation from readers within the specific discipline and general readers.
Appendix \ref{sec:questionnaire} shows the details of these measures and the questionnaire form.

\noindent \textbf{Hyperparameters.}
We list hyperparameters in Appendix \ref{sec:hyper} for brevity.

\subsection{Automatic Evaluation}
The results in Table \ref{tab:auto} show the comparison of different methods.
Recent LLMs of similar scales have shown comparable performance, surpassing the LLaMA-2 introduced in 2023. 
Larger LLMs such as Mixstral-8x7B and Qwen1.5-72B show even better performance, indicating that performance improves as the model scale increases. 
Additionally, the formidable LLM GPT-4 outperforms all the other single LLMs. 
These findings demonstrate their performance on ASJ consistent with the capacity of LLMs.
The fine-tuning methods exhibit competitive performance, slightly superior to prompting the open-source LLMs.
Fine-tuning a model with a larger scale (Qwen1.5-7B) outperforms fine-tuning BART, in line with prompting.

Our collaboration with LLMs has demonstrated significant improvement over previous methods. 
Interestingly, it can be observed that fine-tuning the journalist LLM to warm up does not lead to any noticeable improvement. 
One possible reason may be the decay in the ability to follow instructions after this specialized training.
Nonetheless, these empirical results show the effectiveness of our framework that integrates multiple LLMs.

\begin{figure*}[ht]
  \centering
  \begin{subfigure}{0.328\textwidth}
    \centering
    \includegraphics[height=0.7\columnwidth]{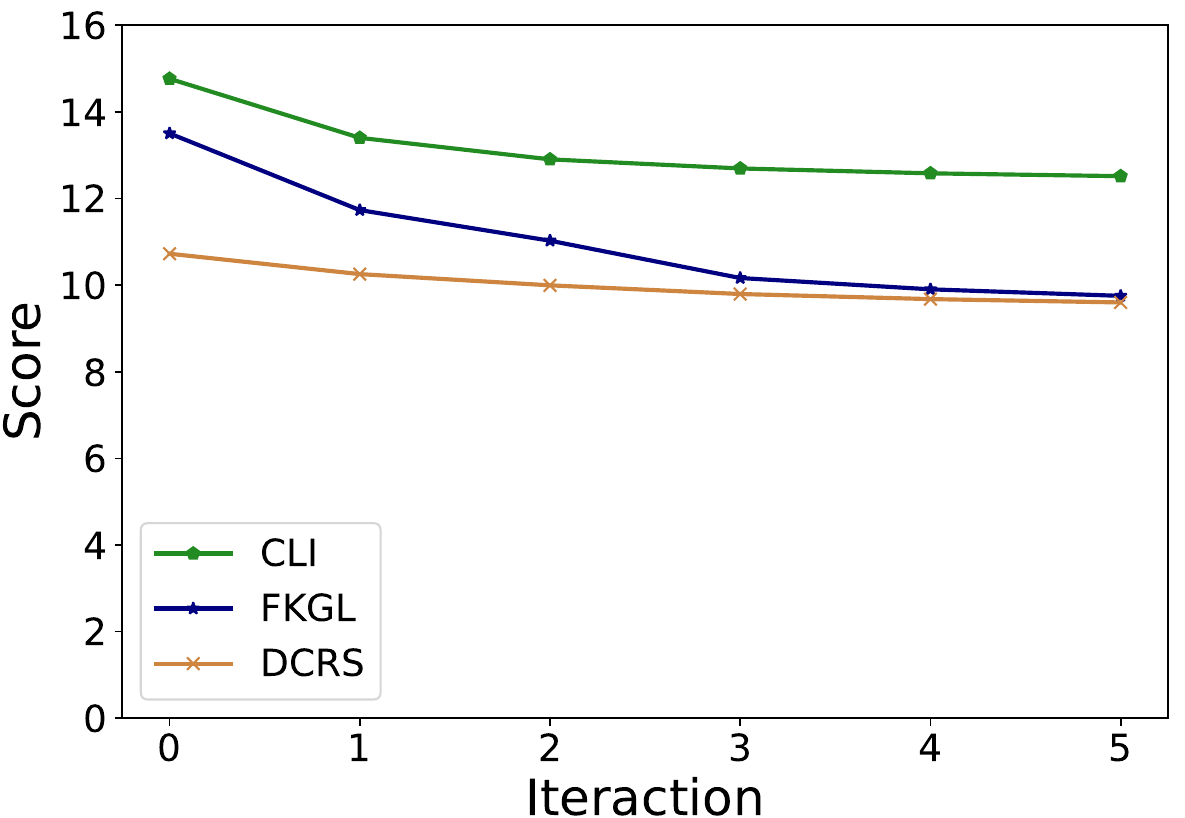}
    \caption{SCITech}
  \end{subfigure}
  \hfill
  \begin{subfigure}{0.329\textwidth}
    \centering
    \includegraphics[height=0.7\columnwidth]{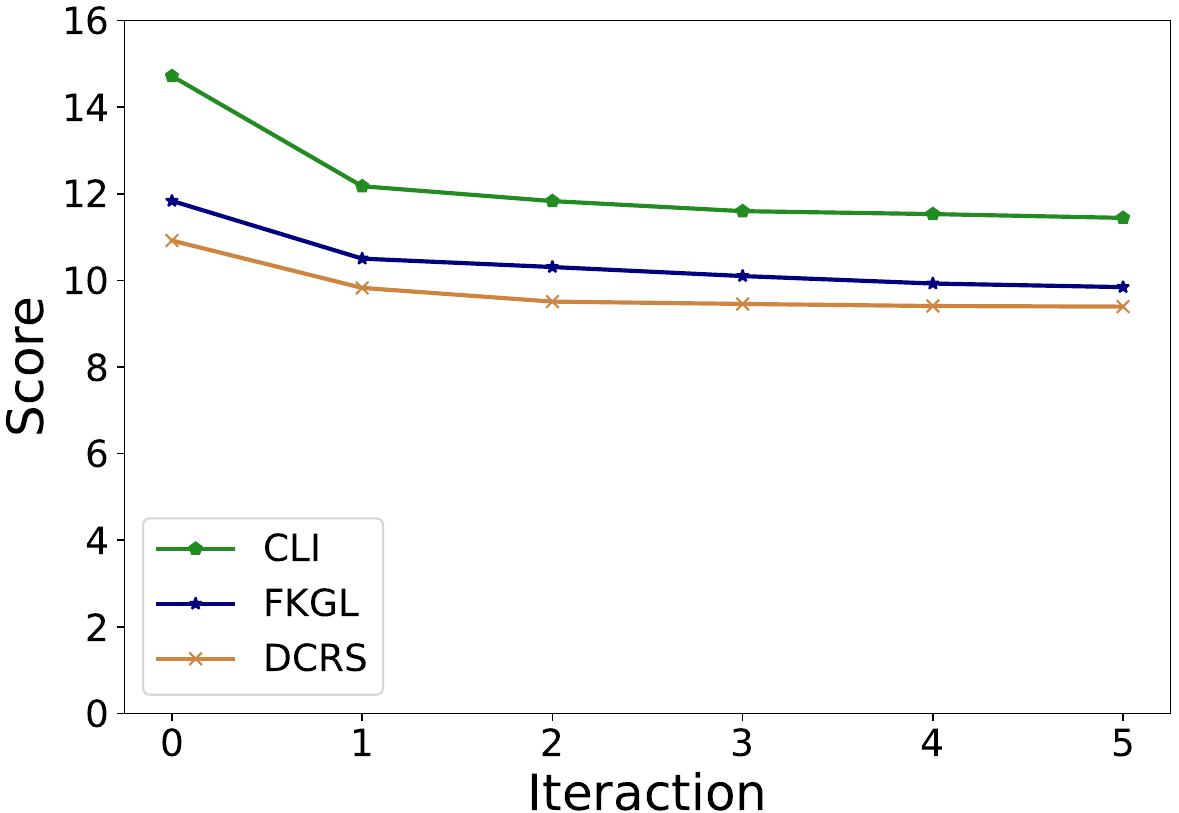}
    \caption{eLife}
  \end{subfigure}
  \hfill
  \begin{subfigure}{0.328\textwidth}
    \centering
    \includegraphics[height=0.7\columnwidth]{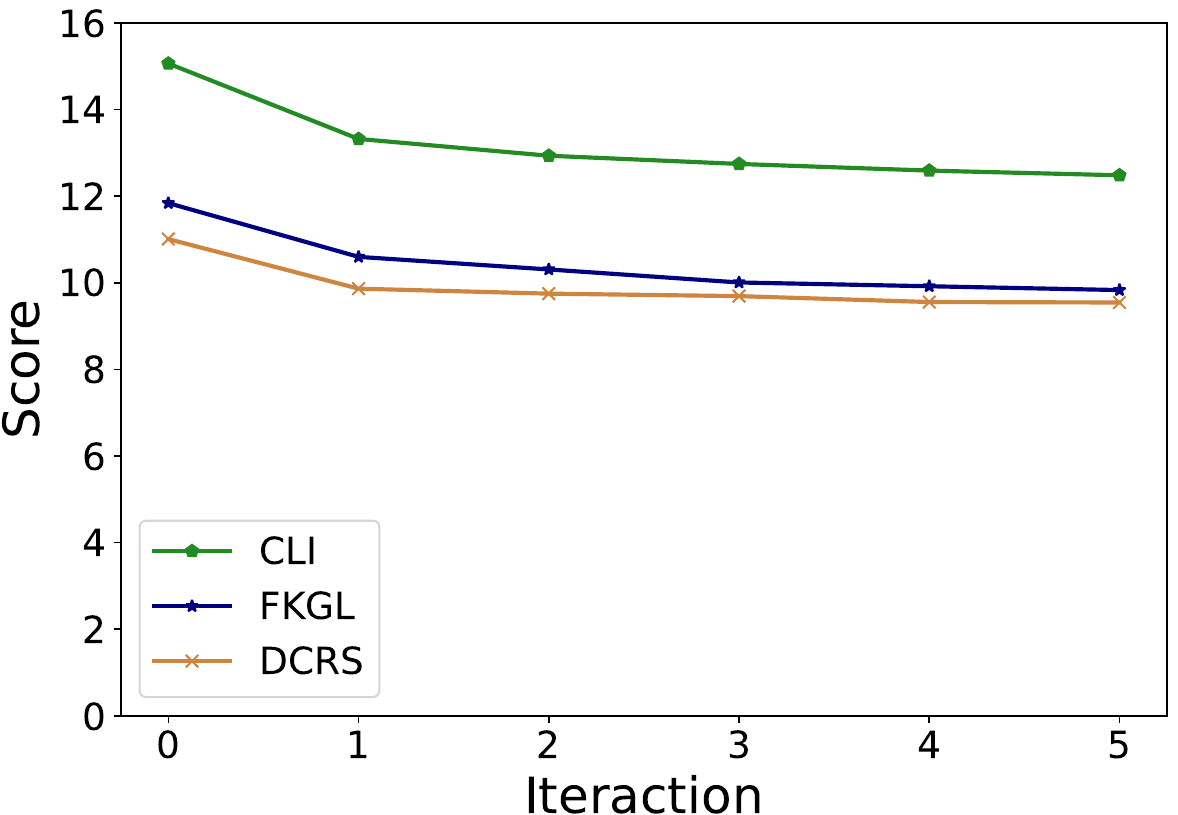}
    \caption{PLOS}
  \end{subfigure}
  \caption{Performance improvement over iterations, with the 0th iteration indicating the initial writing.}
  \label{fig:trend}
\end{figure*}

\begin{table*}[t]
\centering
\small
\begin{tabular}{p{15.4cm}}
\toprule
\textbf{Scientific Paper.} In infectious disease diagnosis, \textit{\textbf{results need to be communicated rapidly to healthcare professionals once testing has been completed so that care pathways can be implemented}}. This represents a particular challenge when testing in remote, low-resource rural communities, in which such diseases often create the largest burden. Here, we report a \textit{\textbf{smartphone-based end-to-end platform for multiplexed DNA diagnosis of malaria}}. The approach uses a low-cost paper-based microfluidic diagnostic test, which is combined with deep learning algorithms for local decision support and blockchain technology for secure data connectivity and management. We validated the approach via field tests in rural Uganda, where it correctly identified more than 98\% of tested cases. Our platform also provides secure geotagged diagnostic information, which creates the possibility of integrating infectious disease data within surveillance frameworks.
\\
\hline
\textbf{Press Release.} Researchers at the U.K.\'s University of Glasgow and Uganda\'s Ministry of Health have developed easy-to-administer blood tests that could help curtail the spread of malaria in sub-Saharan Africa. These "origami" tests can be conducted in the field, with a blood sample taken via fingerprick placed in a channel made from sheets of folded wax paper. The blood is directed into three small chambers of a loop-mediated isothermal amplification (LAMP) machine, which tests for the presence of the mosquito-borne parasite that causes malaria. A smartphone app uses deep learning to ensure the tests are accurate and to monitor community transmission. The results are stored on a blockchain-based ledger to maintain privacy, with anonymized data shared with local authorities. In Uganda, the blood tests were shown to be 98\% accurate.\\
\hline
\textbf{Article Written by Our Method.} In the fight against malaria, \vivid{timely diagnosis is crucial}, especially in rural communities where these diseases often hit hardest. Researchers have come up with a brilliant solution using smartphones! It's a \technical{smartphone-based system that combines a low-cost test with deep learning and blockchain technology}.

First, \vivid{imagine a tiny paper device that can detect malaria DNA right from blood samples}. \technical{This microfluidic test is as simple as using a piece of paper - no fancy lab equipment needed!} Then, the results are analyzed by smart algorithms on your phone. \vivid{These "AI helpers" give instant feedback to healthcare workers without delay.}

All this diagnostic data is stored securely on blockchain. This means information is protected and cannot be tampered with. In Uganda, field tests showed that this system was almost perfect - it correctly identified 98\% of cases!

\vivid{The best part? The platform also records each diagnosis's location through geotags.} This opens up the possibility to \technical{connect local disease data to bigger surveillance networks, helping us track and combat malaria more efficiently.}

So next time you hear about remote villages facing health challenges due to lack of fast diagnosis tools, remember: there's hope with these smartphone-powered innovations!
\\
\bottomrule
\end{tabular}
\caption{An example of the original paper, excerpts from the manually written press release, and the article generated by our method. We highlight statements that we regard \vivid{easy to understand} and \technical{explanations for technical details}.} 
\label{tab:case}
\end{table*}

\begin{table*}[t]
\centering
\small
\begin{tabular}{p{15.4cm}}
\toprule
\textbf{1st Iteration.}
1. The mobile lab on your phone is a portable, paper-based test that uses deep learning algorithms to analyze blood samples for the presence of malaria.
2. The end-to-end system combines a smartphone app, allowing healthcare workers to make informed decisions about treatment without delay.
3. The smartphone app uses \note{deep learning algorithms to interpret the test results}, providing accurate diagnoses over 98\% in field tests.
4. The platform ensures secure data management by using \note{blockchain technology, which records transparently and cannot be tampered with}.
\\
\midrule
\textbf{2nd Iteration.}
1. Smartphones and Blockchain: The article introduces the use of smartphones and \note{blockchain technology in diagnosing diseases like malaria, particularly in remote areas where traditional diagnostic methods may not be available or accessible}.
2. Malaria Diagnosis: The low-cost paper-based microfluidic diagnostic test uses a piece of tiny paper device to detect malaria DNA. It allows rapid detection with the support of AI systems.
3. Low-Cost Test: The AI-based test is referred to as a "smart decision support" system, which uses advanced technology such as artificial intelligence to interpret the results right on the user's device.
4. \note{Artificial Intelligence (AI): The AI acts like a smart detective, quickly identifying potential cases with an accuracy rate of 98\% in real-world tests}, enabling healthcare workers to make informed decisions about treatment without needing fancy labs or waiting too long.
\\
\midrule
\textbf{3rd Iteration.}
1. Smartphone: A device with advanced features such as cameras, sensors, and processors that can perform various tasks, including medical diagnosis.
2. Blood test: A diagnostic procedure where blood is collected from a finger or other body part through a small needle and analyzed for specific signs of malaria (e.g., anemia or fever).
3. \note{AI as personal health assistant: An artificial intelligence system designed to assist individuals in managing their health by providing personalized recommendations based on medical data analysis (e.g., detecting early signs of disease).}
4. Secure digital diary: A digital record containing sensitive information about an individual's health status stored on a secure blockchain network to ensure accuracy and prevent unauthorized access or manipulation.
5. \note{Blockchain technology: A distributed ledger system that allows secure sharing of data across multiple parties without the need for intermediaries or centralized authorities (e.g., storing patient records in hospitals). In this case, it serves as the secure digital ledger for tracking where diseases are spreading in remote areas due to malaria prevalence.}
\\
\bottomrule
\end{tabular}
\caption{Notes taken from the reader, \note{two technical terms and their explanations} are highlighted. The notes become more detailed and comprehensive during the iterative process.} 
\label{tab:notes}
\end{table*}

\begin{table*}[t]
\centering
\small
\begin{tabular}{p{15.4cm}}
\toprule
\textbf{1st Iteration.}
1. Simplify technical terms
2. Break down processes
3. Emphasize benefits
4. Conclusion statement
\\
\midrule
\textbf{2nd Iteration.}
1. \textbf{Emphasize technology}: Instead of listing specific platforms (Ethereum or Hyperledger Fabric), mention that a secure digital ledger is used without specifying which one.
2. \textbf{Simplify medical terms}: For "Plasmodium falciparum," provide a simpler explanation or use a common term like "malaria parasite type."
3. \textbf{Break down process}: Explain how smartphone analysis works in more detail - perhaps by describing how it compares to traditional lab methods.
4. \textbf{Connect to equity}: Highlight how this technology addresses health disparities by providing quick diagnosis in remote areas.
\\
\midrule
\textbf{3rd Iteration.}
1. \textbf{Emphasize simplicity}: For accessibility, rephrase "low-cost paper-based microfluidic diagnostic test" as "affordable, easy-to-use test with a paper strip."
2. \textbf{Explain AI in simpler terms}: Instead of "AI instantly interprets results," say "The smartphone app quickly analyzes the data to give a diagnosis."
3. \textbf{Break down data security}: Highlight that information is stored securely on a phone or cloud server with strong passwords or encryption.
4. \textbf{Quantify success}: Mention that 98\% accuracy rate is exceptional but could be framed as an impressive achievement ("This system detected almost all cases correctly!").
5. \textbf{Cite real-life impact}: Share examples of how this technology has made a difference in remote communities to connect it emotionally with readers.
\\
\bottomrule
\end{tabular}
\caption{Suggestions provided by the editor LLM are becoming increasingly specific over the iterative process.} 
\label{tab:suggest}
\end{table*}

\subsection{Human Evaluation}
For a thorough evaluation, we carry out a human assessment on four representative methods. 
We request that participants assess articles relevant to their fields as well as those in their unfamiliar fields.
The results are reported in Table \ref{tab:human}, categorized into within-field and outside-field articles.
Notably, participants assign lower ratings to articles outside their expertise, possibly due to the inherent comfort and familiarity bias elevating subjective scores.
Despite the evident discrepancies arising from differing reader familiarity contexts, the method comparisons and overarching trends appear consistent across both settings.

Interestingly, all LLM-based methods outperform the plain summaries written by humans, probably because these summaries remain technical, providing a poor reading experience for a wide readership.
The LLM-generated content, targeted to popular science, should be easier for both in-domain and out-of-domain readers to read.
Furthermore, our LLM-collaboration approach surpasses the single Qwen in all dimensions, demonstrating the effectiveness of LLM collaboration.
Most importantly, our LLM collaboration method achieves the highest readability and interest in both within-field and outside-field reading evaluation, with high information conveyance and authenticity close to the most advanced model GPT-4.
Collectively, these findings attest to the potency and effectiveness of our proposed approach.

%% file: src/5-discs.tex
\section{Analysis}
\label{sec:ana}

\subsection{Ablation Study}
We conduct an ablation experiment to validate the effectiveness of each component.
The results have been included in Table \ref{tab:stat} for clarity.
In the ``\textit{Reader: 1.8B $\rightarrow$ 7B}'' setting, we substitute the 1.8B reader model with the 7B version. 
This substitution leads to a minor performance fluctuation.
On the one hand, the 7B model's strong ability gives it greater tolerance for low readability of content, making it harder to highlight writing issues in articles.
On the other hand, it excels in instruction following, enhancing task execution and reducing intermediate errors in the workflow. 
The dynamics between gains and losses render the use of the 7B model as a ``reader'' comparably advantageous at times and disadvantageous at others.
Nevertheless, we recommend using the 1.8B model due to its higher resource efficiency and throughput.

For $-$\textit{Reading Notes}, we eliminate the requirement for the reader LLM to read the article and make notes. 
Instead, we ask the editor LLM to offer suggestions directly.
For $-$\textit{Suggestions}, the editor's advice is omitted, and the journalist revises the article based on the reader LLM's reading.
For $-$\textit{Collaboration}, the journalist revises the article based on the previous writing without any input from the reader or the editor.
As depicted in Table \ref{tab:stat}, our approach exhibits a decrease in performance when each module is removed, underscoring the significance of each module.

\subsection{Trend Analysis}
We provide an in-depth analysis of the performance trajectory over iteration cycles.
The entire writing-reading-suggesting-revising process is carried out for five iterations, with readability score changes depicted in Figure \ref{fig:trend}. 
Our findings reveal a successive and pronounced decline in all reading difficulty metrics over the initial three iterations. 
This pattern demonstrates the efficacy of our iterative revision methodology. 
Following the third iteration, the performance levels off, indicating diminishing improvements from subsequent suggesting and editing efforts.

\subsection{Writing Case Analysis}

To facilitate an intuitive comparison of our method, we present a case study on one writing sample. 
As shown in Table \ref{tab:case}, our method can generate articles that are more readable, with concise and vivid expressions, along with explanations for technical details.
For instance, our generated article states that ``timely diagnosis is crucial'' rather than ``results need to be communicated rapidly to healthcare professionals once testing has been completed so that care pathways can be implemented'', making it more brief and accessible for readers to grasp the research objective.
Moreover, our generated article details that the proposed system is a ``smartphone-based system that combines a low-cost test with deep learning and blockchain technology'' rather than ``smartphone-based end-to-end platform for multiplexed DNA diagnosis of malaria'', enhancing the comprehensibility for a broader audience.

\subsection{Case of Reading Notes}
\label{sec:stu}

We also present a case study on the notes taken by the reader LLM during the first three iterations.
As evidenced in Table \ref{tab:notes}, the notes offer more detailed explanations for technical jargon than those in preceding iterations. 
For example, the technical terms ``AI'' and ``blockchain'' are more thoroughly explained in the third iteration than in the first two iterations. 
This phenomenon suggests that as the readability of writing improves, readers can acquire deeper and more elaborate insights, aligning with human reading behaviours.

\subsection{Suggestions for Revision}
We further examine the advice given by the LLM editor for the revision over the iterative process, as exemplified in Table \ref{tab:suggest}.
In the initial stage, the editor offers general advice. 
Following the revision, the advice becomes more specific and detailed. 
Subsequent rounds of suggestions highlight specific content in the article and recommend revisions based on various aspects. 
In the 3rd iteration, the feedback from the editor becomes more detailed and specific, with specific adjustments from the original version to the revised content.
This revision behaviour is similar to real-world science journalism, showcasing the efficacy of our framework and the social intelligence of LLMs.

%% file: src/2-relat.tex
\section{Related Work}

\noindent \textbf{Automatic Science Journalism.} 
ASJ has gained increasing interest in recent years.
\citet{dangovski2021we} created a parallel corpus and provided a sequence-to-sequence method to generate news summaries from scientific articles.
However, this dataset is not available to the public because of the licensing restriction.
\citet{goldsack2022making} released two corpora, primarily focused on the biomedical and life science domains.
Similarly, they employed a standard sequence-to-sequence model for such tasks.
\citet{cardenas2023don} constructed a dataset in various scientific fields and integrated the discourse structure of papers with their metadata to guide the generation.
These methods of fine-tuning on small models can provide a good match with the reference, but there is still room for improvement in readability.
In this work, we present a novel approach that integrates LLMs acting as agents for enhancing readability.

\noindent \textbf{Large Language Models.} 
% With the emergence of the LLM, advanced models such as GPT-4 \citep{openai2024gpt4}, LLaMA \citep{touvron2023llama}, Claude \citep{anthropic2023claude}, Mistral \citep{jiang2023mistral} and Qwen \citep{bai2023qwen} have shown performance comparable to that of humans in a variety of real-world tasks.
With the emergence of the LLM, advanced models \citep{openai2024gpt4, touvron2023llama, anthropic2023claude, jiang2023mistral, bai2023qwen} have shown performance comparable to that of humans in a variety of real-world tasks.
\citet{park2023generative} leveraged LLMs for social simulation, showing communication and collaboration between LLM-based agents.
\citet{liu2023dynamic} built an LLM-collaboration architecture for enhancing reasoning and code generation tasks.
\citet{qian2023communicative} used different LLM-based agents throughout the development process for software engineering.
This body of work has demonstrated the strong collective intelligence of LLMs.
Inspired by this collection of work, we utilize LLMs as communicative agents, writing, reading, and revising scientific articles to make content accessible to the general audience through a process resembling real-world practice.

%% file: src/6-concl.tex
\section{Conclusion}
This study presents the first examination of LLM collaboration for automatic science journalism aimed at general readers. 
Initially, an LLM functions as a journalist by composing an explanatory article for the general public. 
Subsequently, a smaller LLM, acting as a general audience, reads these articles and takes notes, potentially helping reveal the low readability of the article content. 
An LLM editor assesses the reader's notes and offers suggestions for improvement. 
Following the editor's advice, the journalist revises the manuscript, which is then passed to the reader again to continue the iterative process. 
Extensive experiments are conducted to evaluate the effectiveness of our framework, including both automatic metrics and human evaluation. 
In comparison to directly prompting and fine-tuning various LLMs, our method achieves the highest readability while maintaining high quality.
Additionally, we offer an in-depth analysis, which further validates our method and provides insights into designing LLM-driven ASJ systems.

\section*{Limitations}
We identify the following four limitations of our work.
First, following previous work, we have defined ASJ as the process of transforming a single paper into an article intended for a general audience. 
In practice, a popular science article may encompass multiple studies. 
Therefore, an extension can be to address the challenge of consolidating several papers into a single article.
Second, we have utilized some statistical indexes for automatic assessment, but these statistical measures may miss semantic information. 
LLM-driven evaluation could offer a solution. 
While there remains a gap between LLMs and humans in evaluating text on readability and authenticity, efforts could be made to minimize this gap.
Third, given our exploratory approach in utilizing LLMs for ASJ, we strategically chose abstracts as input to maintain both simplicity and resource efficiency. 
Nevertheless, long-context ASJ is an intriguing task potentially with new challenges to address.
Lastly, all components of our framework are powered by LLMs. 
In addition to our efforts to make each LLM simulate humans, it will be interesting to incorporate genuine human preferences to enhance the generated content.

\section*{Ethics Statement}
The experiments in this study were conducted on publicly available datasets. 
Any information involving privacy was removed. 
All annotators have been properly paid for their efforts.
% Each participant received $\$$70 for completing the human evaluation task, a compensation well above the local income levels.

%% file: src/appdx.tex
\section{Prompts for LLMs}
\label{sec:prompts}
We list all prompts in Table \ref{tab:prompts}.
All prompts follow a similar format.
First, we assign a role to each LLM agent by a sentence.
We then specify the task and background in one or two sentences.
Next, we give each LLM step-by-step instructions.
After that, we input the rules to be followed for each LLM.
Finally, the format of the output is specified as ``markdown'' style to facilitate the extraction and support the information flow among LLMs.
In our preliminary study, this pattern works well in prompting various LLMs, with strong task completion and format adherence.
Algorithm \ref{alg:proc} presents the workflow of our framework.
\begin{algorithm}[t]
\small
\begin{algorithmic}[1]
\Require a scientific paper $\boldsymbol{x}$; journalist $\mathcal{J}$; reader $\mathcal{R}$; editor $\mathcal{S}$; number of iterations $t$.

\State $\boldsymbol{p}_0 \gets \mathcal{J}(\boldsymbol{x})$ \Comment{Initialization}
\For {$i=1$ to $t$}
    \State $\boldsymbol{r} \gets \mathcal{R}\left(\boldsymbol{p}_{i-1}\right)$   \Comment{Reader's notes}
    \State $\boldsymbol{s} \gets \mathcal{S}\left(\boldsymbol{p_{i-1}, r}\right)$  \Comment{Editor's suggestions}
    % \State $\boldsymbol{p}_i \gets \mathcal{J}\left(\boldsymbol{x, s}\right)$ \Comment{Revision}
    \State $\boldsymbol{p}_i \gets \mathcal{J}\left(\boldsymbol{x, p_{i-1}, s}\right)$ \Comment{Revision}
    
\EndFor \\
\Return $\boldsymbol{p}_t$
\end{algorithmic}
\caption{LLM-Collaboration for ASJ}
\label{alg:proc}
\end{algorithm}

\section{Failed Attempts}
\label{sec:fail}
This appendix outlines our unsuccessful attempts. 
We hope that it will save time for other researchers.
First, as a commonly used mechanism in LLM agents, reflection can support iterative enhancement by consolidating prior experiences \citep{yao2022react, park2023generative, yang2023failures}. 
In our pilot experiments, however, this approach did not succeed in refining the journalist's writing.
One potential explanation could be that the varied nature of journalistic content necessitates more targeted revision instructions, whereas summarizing writing experiences results in general writing guidance.

Within our framework, an LLM acts as an audience, reading the written article and taking notes.
How about having this audience perform the reading comprehension task instead of taking notes?
Intuitively, it can also demonstrate the reader's understanding, and then induce the content complexity.
Preliminary investigations, however, revealed that this approach yields lesser efficacy compared to the note-taking strategy.
One possible reason might be that the quality of question generation greatly affects the efficiency of the whole workflow.
Besides, asking a fixed number of questions narrows the expanse of textual exploration, thereby constricting the comprehensive perception of content complexity.

\section{Dataset and Hyperparameters}
\label{sec:hyper}
Table \ref{tab:stat} shows some statistics of the three datasets used in our experiments.
\emph{SCITech(News)} is released by \citet{cardenas2023don}, who gathered press releases from ACM Technews as well as their source articles from various publishers, involving fields of computer science, engineering, astrophysics, biology, and others. 
\emph{eLife} is an open-access journal that focuses on biomedical and life sciences. 
\citet{goldsack2022making} collected some eLife articles as well as digests written by expert editors based on both the article itself and questions answered by the author. 
Similarly, \emph{PLOS} hosts journals across areas of science and medicine. 
Some of these articles, also collected by \citet{goldsack2022making}, come with the author's summary. 
For resource saving, the original paper's abstract serves as the scientific content input.
\begin{table}[t]
    \small
    \centering
    \begin{tabular}{lrrrrr}
        \toprule
       Statistic & \textbf{SCITech} & \textbf{eLife} & \textbf{PLOS} \\
        \midrule
        \# $\text{pairs}$ & 2431 & 4828 & 27525 \\
        \# $\text{words}^\text{ori}$ & 216.8 & 166.3 & 268.3\\
        \# $\text{sentences}^\text{ori}$ & 5.7 & 6.8 & 10.2 \\
        \# $\text{words}^\text{pln}$ & 176.1 & 347.6 & 175.6 \\
        \# $\text{sentences}^\text{pln}$ & 7.9 & 15.7 & 7.8\\
        \bottomrule
    \end{tabular}
    \caption{Statistics of benchmark datasets. Words and sentences are average values. The ``ori'' superscript indicates abstracts of original papers, and ``pln'' represents plain summaries written by authors or journalists.}
    \label{tab:stat}
\end{table}

Our local LLM service runs on a machine with eight GTX 4090 GPUs.
We utilize the Huggingface platform \citep{wolf2019huggingface} for downloading and loading checkpoints.
For rapid inference and memory efficiency, we utilize the vLLM library\footnote{\url{https://docs.vllm.ai}} to develop API services.
We deploy agents from the Qwen-1.5 series, for their good performance and diverse model scales \cite{bai2023qwen}.
In particular, Qwen-1.5-7B is employed for the steps of writing, providing suggestions, and revision, whereas Qwen-1.5-1.8B serves as the reader for taking notes.
To improve memory efficiency, we implement activation-aware weight quantization (AWQ, \citealt{lin2023awq}) for model quantization.
For fine-tuning, we utilize LoRA \citep{hu2021lora} with Llama-Factory \citep{zheng2024llamafactory}, adopting the default setting with the number of the epochs set to 10.
We use the default temperature setting and empirically set top\_p to 0.4, frequency penalty and repetition penalty to 1, ensuring the stability of the LLMs' output while retaining diversity. 
We iterate five times and empirically select the output from the third iteration as the final result.
The maximum number of tokens in the model output is 4096.

\section{Details of Human Evaluation}
\label{sec:questionnaire}
\begin{figure*}[t]
    \centering
    \includegraphics[width=1\textwidth]{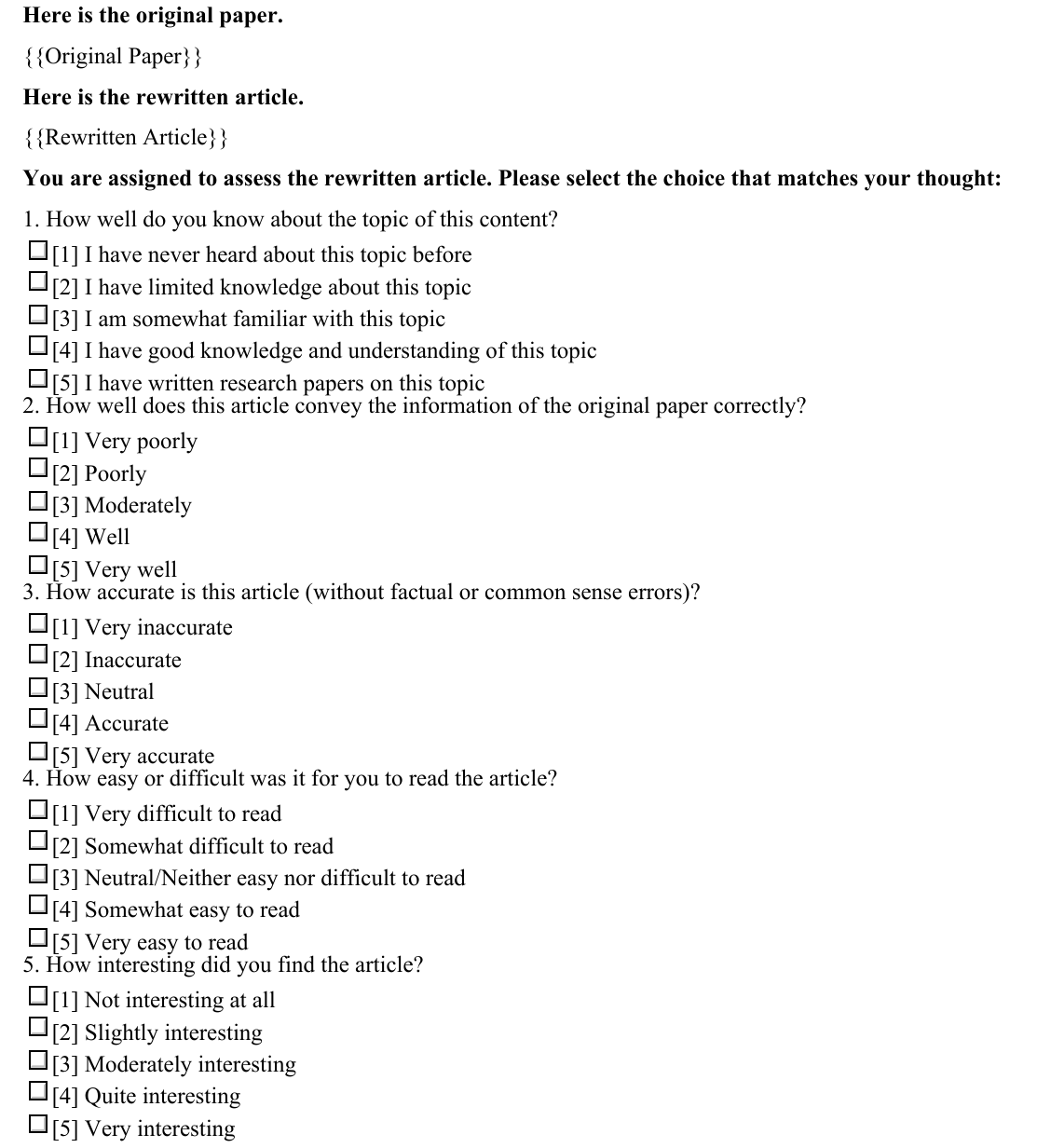}
    \caption{The questionnaire for participants to evaluate articles.}
    \label{fig:question}
\end{figure*}

We create a questionnaire for human evaluation using a 1-5 Likert scale, as shown in Figure \ref{fig:question}. 
All participants were informed that their assessments would be used for research purposes.
We utilize Label Studio \citep{Label-Studio} to construct the annotation platform. 
Initially, participants indicate their familiarity with the given topic.
They are then tasked with answering 4 questions related to Readability, Information Conveyance, Authenticity, and Interestingness.
Readability assesses how easily the article can be read, serving as a supplementary and further validation of automated evaluation. 
Information conveyance determines if the rewritten content accurately and comprehensively conveys the information from the original paper. 
Similarly, authenticity assesses the correctness of the content. 
A high-quality article should contain minimal factual or common sense errors to avoid misleading readers. 
The level of interest is also a crucial factor; content of high appeal will attract more readers.

% \section{Discussion on Metrics}
% \label{sec:disc}

% This is an appendix.

\section{Use of AI Assistants}
We use ChatGPT for correcting grammar and improving expressions in this manuscript.

\begin{table*}[t]
\centering
\small
\begin{tabular}{p{15.56cm}}
\toprule
\textbf{Journalist.}
You are a science journalist for general audiences. Given a paper's summary, you are assigned to rewrite it into a short understandable article for general audiences.\\
Follow the rules strictly: \\
- Keep short yet informative.\\
- The output format:\\
\#\# Article\\
...\\
\midrule
\textbf{Reader.}
You are a general reader. Given a popular science article, please read it carefully and take some notes.\\
Please take the following steps:\\
1. First, extract all technical terms with their context from the article.\\
2. Then, explain the technical terms based on their context.\\
Follow the rules strictly:\\
- Extraction should mention the specific location of each technical term in the article.\\
- Explanation should be first extracted from the article; if not found, it can be some common-sense or specialized knowledge.\\
- Extraction and explanation should be in points, like "1...2...3...".\\
- The output format:\\
\#\#\# Extraction\\
1. ...\\
2. ...\\
...\\
\#\#\# Explanation\\
1. ...\\
2. ...\\
\midrule
\textbf{Editor.}
You are a senior editor. Here are a scientific paper summary, and a short popular science article. A general reader has read the science article and takes some notes.\\
Please take the following steps:\\
1. First, evaluate the **reader's notes** based on these factors: content accuracy, lexical and technical complexity, and information conveyance (from the original content). \\
2. Then, based on the above evaluation, list some brief yet informative writing advice that may benefit the popular science article, to make the article easier for general readers without specialized knowledge to read and understand. Specifically, the advice should benefit these factors of the article:\\
    a) Content Accuracy: The factual correctness, scientific validity, and absence of errors in the general popular science article. \\
    b) Accessibility: Higher accessibility means less technical, more readable and interestingly, etc.
    c) Information Conveyance: How effectively key information from the original paper is transferred to the popular science article.\\
Follow the rules strictly: \\
- Evaluation and advice sections should be in points, like "1...2...3...".\\
- Each advice should not go beyond the fact of original paper, but can be some common-sense or specialized knowledge.\\
- Each advice should be targeted at one specific aspect of the article.\\
- Don't suggest visualization, references and links.\\
- Suggest explanations rather than content additions.\\
- The output format: \\
\#\# Evaluation for reader's notes\\
- Content accuracy of reader's notes: ...\\
- Lexical and technical complexity of reader's notes: ...\\
- Information conveyance of reader's notes: ...\\
\#\# Advice\\
1. ...\\
2. ...\\
\midrule
\textbf{Revision}
You are a science journalist for general audiences. Given the paper summary and a short summary of the popular science article, you are assigned to rewrite the popular science summary for general audiences, who have no specialized knowledge on this field. There are some writing advice.\\
Please take the following steps: \\
1. Choose and refine the most relevant and suitable advice for writing improvement.\\
2. Then, based on the refined advice and the paper summary, rewrite the popular science article.\\ 
Follow the rules strictly: \\
- Keep short yet informative.\\
- Don't include visualization, references and links.\\
- Revision must not go beyond the fact of original paper, but can be with some additional common-sense or professional knowledge for explanation.\\
- The output format:\\
\#\# Improvement\\
...\\
\#\# Revised Article\\
...\\
\bottomrule
\end{tabular}
\caption{Prompts for each LLM agent.} 
\label{tab:prompts}
\end{table*}

